\documentclass[runningheads]{llncs}
\usepackage{graphicx}
\usepackage{url}
\usepackage{latexsym}
\usepackage{paralist}
\usepackage[shortlabels]{enumitem}
\usepackage{graphicx}
\usepackage{xcolor}
\usepackage{multirow}
\usepackage{paralist}
\usepackage{amsmath}
\usepackage{hyperref}
\usepackage{comment}
\begin{document}

\title{Drug and Disease Interpretation Learning with Biomedical Entity Representation Transformer}
\titlerunning{Drug and Disease Representation Transformer}

\author{Zulfat Miftahutdinov\orcidID{0000-0002-8467-4824} \and
Artur Kadurin\orcidID{0000-0003-1482-9365} \and
Roman Kudrin\orcidID{0000-0002-9741-0043} \and
Elena Tutubalina\orcidID{0000-0001-7936-0284}
}
\authorrunning{Z. Miftahutdinov et al.}
%
\institute{Insilico Medicine Hong Kong, Pak Shek Kok, Hong Kong \\
\email{zulfat,artur,kudrin,elena@insilico.com}\\
\url{https://insilico.com/}
}

\maketitle              
\begin{abstract}
Concept normalization in free-form texts is a crucial step in every text-mining pipeline. Neural architectures based on Bidirectional Encoder Representations from Transformers (BERT) have achieved state-of-the-art results in the biomedical domain. In the context of drug discovery and development, clinical trials are necessary to establish the efficacy and safety of drugs. We investigate the effectiveness of transferring concept normalization from the general biomedical domain to the clinical trials domain in a zero-shot setting with an absence of labeled data. We propose a simple and effective two-stage neural approach based on fine-tuned BERT architectures. In the first stage, we train a metric learning model that optimizes relative similarity of mentions and concepts via triplet loss. The model is trained on available labeled corpora of scientific abstracts to obtain vector embeddings of concept names and entity mentions from texts. In the second stage, we find the closest concept name representation in an embedding space to a given clinical mention. We evaluated several models, including state-of-the-art architectures, on a dataset of abstracts and a real-world dataset of trial records with interventions and conditions mapped to drug and disease terminologies. Extensive experiments validate the effectiveness of our approach in knowledge transfer from the scientific literature to clinical trials. 

\end{abstract}

\section{Introduction}
The emerging use of neural network architectures in the early-stage of drug discovery has recently resulted in several breakthroughs~\cite{zhavoronkov2019deep,ivanenkov2019identification}. 
Later stages of drug development are much more conservative due to the complicated process of clinical trials. The use of state-of-the-art neural network approaches in clinical trials could dramatically speed up the overall drug development process and increase its success rate, thus saving lives.

Clinical trial registers (e.g., \href{https://clinicaltrials.gov/}{ClinicalTrials.gov}) contain vast amounts of structured information on how standardized interventions work in a clinical setting. Despite the existing structure, these registers remain very difficult to harmonize with drug and disease databases using current techniques. This very often results in substantial information losses. The primary cause for  this inaccurate harmonization is that in a clinical trial record diseases and interventions are not described with a centralized standardized taxonomy but with a free text. The automatic natural language processing (NLP) methods are promising approaches for the semantic annotation of large volumes of clinical records and for the integration and standardization of biomedical entity mentions to formal concepts. In biomedical research and healthcare, the entity linking problem is known as medical concept normalization (MCN). A source as a knowledge base (KB) contains further information about the concept, such as its preferred name and synonyms, pharmacological profile, and its relationships with other concepts.


Neural architectures have been widely used in recent state-of-the-art models for MCN from user reviews and social media texts \cite{leaman2016taggerone,zhao2019neural,zhu2019latte,li2017cnn,miftahutdinov2019deep,tutubalina2018medical}.  
These studies mostly share limitations regarding a supervised classification framework: binary or multiclass classifiers are trained on a dataset with a narrow subsample of concepts from a specific terminology. In particular, recent models \cite{leaman2016taggerone,zhao2019neural,zhu2019latte} learn a scoring function measuring the similarity between an entity mention and a concept. The difficulty with these methods is that it is not possible to extract representations describing mentions and concepts separately. In this setup, to retrieve concepts from a particular terminology for a given entity mention, we have to compute all the similarities through the ranking function and sort these scores in descending order. This is impractical if we need to process large corpora of free-form clinical trials, scientific literature, patents in days.

Inspired by metric learning~\cite{huang2013learning,schroff2015facenet,hoffer2015deep}, its usage for multimodal and sentence representation learning~\cite{liu2017learning,reimers2019sentence}, negative sampling~\cite{mikolov2013distributed}, and Bidirectional Encoder Representations from Transformers (BERT) \cite{devlin2018bert}, we present a BERT-based neural model for medical concept normalization that directly optimizes the BioBERT representations~\cite{lee2019biobert} of entity mentions and concept names itself, rather than classification or ranking layer. We use triplets of free-form entity mention, positive concept names, and randomly sampled concept names as negative examples to train our model. In this work, we consider the zero-shot scenario because it is often the case in the biomedical domain, where there are dozens of concept categories and terminologies. We trained models on annotated pairs of disease or chemical mentions with the corresponding concepts 
and evaluated on a novel dataset of condition and intervention concepts from clinical trials. 

The contributions of this paper can be summarized as follows:
\begin{enumerate}
\item We develop a simple and effective model that uses metric learning and negative sampling to obtain entity and concept embeddings. These embeddings were utilized for knowledge transfer between different terminologies. We explore several strategies to select positive and negative samples.
\item We perform extensive experiments of several BERT-based models on a newly annotated dataset of clinical trials in two setups, where each mention is associated with one or more concepts (in-KB) or zero (out-of-KB). 
\end{enumerate}


\section{Related Work}
Our work most closely relates to research in information extraction and semantic textual similarity by directly linking a set of entity mentions and a large set of medical concept names using triplet structures to derive embeddings of entity mentions and concept names that can be compared using semantic similarity. Entity linking of mentions to entries in a knowledge base (KB) is a well-studied area; see a good survey \cite{shen2014entity}. Research studies in this area assume that there is one knowledge base, such as Wikipedia or Freebase. The KB contains rich text descriptions (from an entity page, for example), hyperlink statistics, and meta-data. This assumption holds for the general domain, but not for the biomedical domain, where diverse terminologies exist for numerous purposes.

\subsection{Medical Concept Normalization}
Medical concept normalization is usually formulated as a classification or ranking problem with a wide variety of features – syntactic and morphological parsing, dictionaries of medical concepts and their synonyms, distances between raw entity mentions and formal concept names in terms of TF-IDF or \text{word2vec} representations \cite{aronson2001effective,ghiasvand2014uwm,van2016erasmus,leaman2016taggerone,dermouche2016ecstra}. MetaMap is one of the most well-known knowledge-based systems for mapping texts to concepts from Unified Medical Language System (UMLS)~\cite{bodenreider2004unified} developed by the US National Library of Medicine (NLM)~\cite{aronson2001effective}. This system is based on a linguistic approach using lexical lookup and variants by associating a score with phrases in a sentence. The NLM provides automatic indexing of clinical trials to Medical Subject Headings (MeSH)~\cite{coletti2001medical} via the Medical Text Indexer (MTI)~\cite{mork2013nlm} based on MetaMap. MTI achieves an F1 measure around 0.55 on the indexing of PubMed abstracts. The most popular open-source supervised system maintained by the NLM is TaggerOne \cite{leaman2016taggerone}. TaggerOne utilizes semi-Markov models with features and dictionaries to jointly perform entity extraction and normalization tasks. 

The works that are the closest to ours and consider synonyms during entity and concept representation learning is Biomedical Named Encoder (BNE)~\cite{phan2019robust} and BioSyn~\cite{sung2020biomedical}. Sung et al. proposed a BioBERT-based model named BioSyn that maximizes the probability of all synonym representations in the top 20 candidates~\cite{sung2020biomedical}. BioSyn uses a combination of two scores, sparse and dense, as a similarity function. Sparse scores are calculated on character-level TF-IDF representations to encode morphological information of given strings. Dense scores are defined by the similarity between \texttt{CLS} tokens of a single vector of input in BioBERT. This model achieves state-of-the-art results in disease and chemical mapping over previous works~\cite{leaman2016taggerone,wright2019normco,phan2019robust}. Phan et al. presented an encoding framework with new context, concept, and synonym-based objectives~\cite{phan2019robust}. Synonym-based objective enforces similar representations between synonymous names, while concept-based objective pulls the name’s representations closer to its concept’s centroid. 
However, ranking on these embeddings shows worse results on three sets than TaggerOne. 

Our work differs from the studies discussed above in the following important aspects. First, none of these methods have been applied to free-form descriptions of conditions and interventions from clinical trials. Second, evaluation strategies in the mentioned papers are based on train/test splits provided by datasets' authors. We follow the recent \emph{refined} evaluation strategy from \cite{tutubalina2020fair} on the creation of test sets without duplicates or exact overlaps between the train and test sets. Finally, our dataset includes entity mentions for both in-KB and out-of-KB linking. 


\subsection{NLP in Clinical Trials Research}
While the majority of biomedical research on information extraction primarily focused on scientific literature~\cite{huang2015community}, much less work had been used NLP methods to conduct curation of clinical trial records' fields to advance downstream tasks \cite{gayvert2016data,brown2017standard,atal2016automatic,boland2013feasibility,hao2014clustering,sen2018representativeness}. 
Gayvert et al.~\cite{gayvert2016data} proposed an approach for the prediction of the likelihood of toxicity in clinical trials. They selected 108 clinical trials of any phase that were annotated as having failed for toxicity reasons. Then intervention names of each trial were manually mapped to DrugBank~\cite{wishart2006drugbank} concepts to collect molecular weight, polar surface area, and other compounds' properties. 
In \cite{atal2016automatic}, Atal et al. developed a knowledge-based approach to classify entity mentions to disease categories from a Global Burden of Diseases (GBD) cause list. The proposed method uses MetaMap to extract UMLS concepts from trial fields (health condition, public title, and scientific title), link UMLS concepts with ICD10 codes, and classify ICD10 codes to candidate GBD categories. The developed classifier identified GBD categories for 78\% of the trials.
Li and Lu~\cite{li2012systematic} identified clinical pharmacogenomics (PGx) information from clinical trial records based on dictionaries from a pharmacogenomics knowledge base PharmGKB. 
 Previous studies on clinical trial records, however, have not analyzed the performance of linking of clinical trials to disease and drug concepts, but rather across eligibility criteria (e.g., patient’s demographic, disease category)~\cite{boland2013feasibility,hao2014clustering,sen2018representativeness,atal2016automatic,leveling2017patient}.
\section{Dataset of Clinical Trials}
NLM maintains a clinical trial registry data bank ClinicalTrials.gov\footnote{\urlstyle{tt}\url{https://clinicaltrials.gov/}} that contains over 340,000 trials from 214 countries. This database includes comprehensive scientific and clinical investigations in biomedicine~\cite{gill2016emerging}. Each trial record provides information about a trial’s title, purpose, description, condition, intervention, eligibility, sponsors, etc. Most information from records is described in natural language. In our study, we use publicly available American Association of Clinical Trials (AACT) Database\footnote{\urlstyle{tt}\url{https://www.ctti-clinicaltrials.org/aact-database}}, v. 20200201. 

Since there is no off-the-shelf manually annotated dataset for biomedical concept normalization of clinical trials, we built one by selecting 500 trials using the following criteria:
\begin{enumerate}
    \item A type of clinical study is an interventional study.  Participants of interventional studies receive intervention/treatment so that researchers can evaluate the effects of the interventions on biomedical or health-related outcomes~\cite{medicine2015sharing}.
    \item Phase of clinical study is defined by U.S. Food and Drug Administration (FDA). There are five phases: Early Phase 1, Phase 1, Phase 2, Phase 3, and Phase 4. 
    \item Clinical study is associated with one or more interventions of the following types: Biological, Combination Product, Drug.
\end{enumerate}
As a drug terminology source, we use an internal knowledge base that contains 15,532 concept unique identifiers (CUIs), including small molecule drugs, biologics, nutraceuticals, and experimental drugs. As a condition terminology source, we use MeSH v. 20200101. 500 selected trials contain 1075 and 819 entries in the `Intervention' and  `Condition' fields respectively. Two annotators with a background in bioinformatics manually annotated each entry. The calculated inter-annotator agreement (IAA) using Kappa was 92.32\% for the entire dataset. The disagreement was resolved through mutual consent.  

Statistics of annotated texts are summarized in Table \ref{tab:ctstats}. 794 out of 1075 non-unique mentions (73.9\%) were mapped to one or more drug concepts. 838 (80\%) of lower-cased interventions are unique. 804 out of 819 non-unique mentions (98.2\%) were mapped to one or more concepts, while there are 638 (78\%) lower-cased unique mentions. Interestingly, MeSH concepts linked to conditions belong to several MeSH categories including Diseases [C], Psychiatry and Psychology [F], and Analytical, Diagnostic and Therapeutic Techniques, and Equipment [E]. We note that NLM provided automatically assigned MeSH terms to trials' interventions. 716 out of 1075 entries (66.6\%) were mapped to MeSH terms. Our analysis revealed that mapping from NLM does not include investigational drugs, which are essential for developing new pharmaceutical drugs. Table \ref{tab:sample} contains a sample of annotated texts.

\begin{table}[t]
\centering
\caption{Statistics of annotated texts.} 
\label{tab:ctstats}
\begin{tabular}{|l|l|p{2cm}|l|p{2.5cm}|}
\hline
\textbf{Mention} & \textbf{\#texts} & \textbf{\#texts with CUIs} & \textbf{\#unique texts} & \textbf{\#unique texts with CUIs}\\
\hline
\multicolumn{5}{|c|}{Intervention types} \\
\hline
Drug                & 850 & 693 & 671 & 585 \\
Biological          & 118 & 90  & 102 & 79  \\
Other               & 57  & 4   & 27  & 4   \\
Procedure           & 19  & 1   & 16  & 1   \\
Radiation           & 11  & 0   & 9   & 0   \\
Device              & 11  & 1   & 11  & 1   \\
Combination Product & 5   & 3   & 5   & 3   \\
Dietary Supplement  & 2   & 2   & 2   & 2   \\
Diagnostic Test     & 1   & 0   & 1   & 0   \\
Behavioral          & 1   & 0   & 1   & 0   \\
\hline
\multicolumn{5}{|c|}{Total} \\
\hline
Intervention & 1075 & 794 & 838 & 671\\
Condition & 819 & 804 & 638 & 638\\
\hline
\end{tabular}
\end{table}

\begin{table}[t]
\centering
\caption{Sample of manually annotated trials' texts.} 
\label{tab:sample}
\begin{tabular}{|p{2.2cm}|p{5cm}|p{4.5cm}|}
\hline
\textbf{NCT / Type} & \textbf{Text} & \textbf{Concept} \\
\hline
\multicolumn{3}{|c|}{Intervention (with DrugBank CUIs)} \\
\hline
NCT00559975 / Biological	& Adjuvanted influenza vaccine combine with CpG7909	& Agatolimod sodium (DB15018) \\ \hline
NCT01575756 / Biological	&Haemocomplettan® P or RiaSTAPTM& Fibrinogen human (DB09222)\\ \hline
NCT00081484 / Drug	&epoetin alfa or beta & Erythropoietin (DB00016)\\ \hline
NCT03375593 / Drug	&Ibuprofen 600 mg tab & Ibuprofen (DB01050)\\ \hline
NCT01170442 / Drug	&vitamin D3 5000 IU & Calcitriol (DB00136)\\ \hline
NCT02493335 / Drug	&Placebo orodispersible tablet twice daily & \textit{nil (no concept)} \\
\hline
\multicolumn{3}{|c|}{Condition (with MeSH CUIs)} \\
\hline
NCT02009605	& Squamous Cell Carcinoma of Lung & Carcinoma, Non-Small-Cell Lung	(D002289) \\
NCT04169763 &	Stage IIIC Vulvar Cancer AJCC v8 & Vulvar Neoplasms	(D014846) \\
\hline

\end{tabular}
\end{table}

\section{Model}
In this section, we present a neural model for Drug and disease Interpretation Learning with Biomedical Entity Representation Transformer (\texttt{DILBERT}). We address MCN as a retrieval task by fine-tuning the BERT-based network using metric learning~\cite{huang2013learning,schroff2015facenet,hoffer2015deep}, negative sampling~\cite{mikolov2013distributed}, specifically, triplet constraints. This idea was successfully applied to learn multimodal embeddings~\cite{wu2013online,liu2017learning} and recent sentence embeddings via a sentence-BERT model~\cite{reimers2019sentence}. Compared to a pair of independent sentences or images, two concept names can have relationships as synonyms, hypernyms, hyponyms, etc., that we consider during the training phase to facilitate the concept ranking task at the retrieval phase.

Let us first recall two terms: \textit{concept} and \textit{concept name}. Following the UMLS Glossary~\cite{umlsglossary}, the concept is the fundamental unit of meaning in terminology. It represents a single meaning in any way, whether formal or casual, verbose or abbreviated. Every concept is assigned a unique identifier (CUI). A concept consists of atoms, which are the smallest units of naming. All of the atoms within a concept are synonymous. The concept name is a string chosen to represent the concept as a whole. It is linked to atoms. Formally, the medical concept normalization task aims to assign each entity mention $m$ a CUI (or predicts that there is no corresponding concept). 


\paragraph{Architecture}
Following denotations proposed by \cite{humeau2019poly}, we encode both entity mention $m$ and candidate concept name $c$ into vectors:
\begin{equation}
y_m = red(T(m)); y_c = red(T(c))
\end{equation}
where $T$ is the transformer that is allowed to update during fine-tuning. $red(\cdot)$ is a function that reduces that sequence of vectors into one vector. There are two main ways of reducing the output into one representation via $red(\cdot)$: choose the first output of T (corresponding to the token CLS) or compute the elementwise average over all output vectors to obtain a fixed-size vector. As a pretrained transformer model, we use BioBERT base v1.1. \cite{lee2019biobert}

\paragraph{Scoring}
The score of a candidate $c_i$ for an entity mention $m$ is given by a distance metric, e.g. Euclidean distance:
\begin{equation}
s(m, c_i) = ||y_m - y_{c_i}||
\end{equation}

A noteworthy aspect of the proposed model is its scope: by design, it aims at the cross-terminology mapping of entity mentions to a given lexicon without additional re-training. This approach allows for fast, real-time inference, as all concept names from a terminology can be cached. This is a necessary requirement for processing biomedical documents of different subdomains such as clinical trials, scientific literature and drug labels.

\paragraph{Optimization}
The network is trained using a triplet objective function. Given a user-generated entity mention $m$, a positive concept name $c_g$ and a negative concept name $c_n$, triplet loss tunes the network such that the distance between $m$ and $c_g$ is smaller than the distance between $m$ and $c_n$. Mathematically, we minimize the following loss function:
\begin{equation}
    max(s(m, c_g) - s(m, c_n) + \epsilon, 0)
\end{equation}

where $\epsilon$ is margin that ensures that $c_g$ is at least $\epsilon$ closer to $m$ than $c_n$. As a scoring metric, we use Euclidean distance or cosine similarity and we set $\epsilon = 1$ in our experiments. 

\paragraph{Positive and Negative Sampling}
Suppose that a pair of the entity mention with the corresponding CUI is given as well as the vocabulary. For positive examples, vocabulary is restricted to the concepts that have the same CUI as a mention. Multiple positive concept names could be explained by the presence of synonyms in the vocabulary. Negative sampling~\cite{mikolov2013distributed} uses the rest part of the vocabulary. We explore several strategies to select positive and negative samples for a training pair (entity mention, CUI):
\begin{enumerate}
    \item \textbf{random sampling}: we sample several concept names with the same CUI as positive examples and random negatives from the rest of the vocabulary; 
    \item \textbf{random + parents}: we sample $k$ concept names from the concept's parents in addition to positive and negative names gathered with the random sampling strategy;
    \item \textbf{re-sampling}: using a model trained with random sampling, we identify positives and \textit{hard} negatives via the following steps: (i) encode all mentions and concept names found in training pairs using the current model, (ii) select positives with the same CUI, which are closest to a mention, (iii) for each mention, retrieve the most similar $k$ concept names (i.e., its nearest neighbors) and select all names that are ranked above the correct one for the mention as negative examples. We follow this strategy from~\cite{gillick2019learning};
    \item \textbf{re-sampling + siblings}: we modify the re-sampling strategy by using $k$ concept names from the concept's siblings as negatives.
\end{enumerate}

\paragraph{Inference}
At inference time, the representation
for all concept names can be precomputed and cached. The inference task is then reduced to finding the closest concept name representation to entity mention representation in a common embedding space. 
\section{Experiments}

We evaluate our model \texttt{DILBERT} and compare it to the state-of-the-art methods using (i) a publicly available benchmark BioCreative V CDR Disease \& Chemical \cite{li2016biocreative}, (ii) our dataset of clinical trials named CT Condition \& Intervention.
The statistics of the two datasets are summarized in Table \ref{tab:stats}.

\subsection{Datasets}



BioCreative V CDR \cite{li2016biocreative} introduces a challenging task for the extraction of chemical-disease relations (CDR) from PubMed abstracts. Disease and chemical mentions are linked to the MEDIC~\cite{davis2012medic} and CTD~\cite{davis2019comparative} dictionaries, respectively. We utilize the CTD chemical dictionary (v. November 4, 2019) that consists of pf 171,203 CUIs and 407,247 synonyms, and the MEDIC lexicon (v. July 6, 2012) that contains 11,915 CUIs and 71,923 synonyms.

According to the BioCreative V CDR annotation guidelines, the annotators used two MeSH branches to annotate entities: (i) ``Diseases'' [C], including
signs and symptoms, (ii) ``Drugs and Chemicals'' [D]. 
The terms ``drugs'' and ``chemicals'' are often used interchangeably.
Annotators annotated chemical nouns convertible to single atoms, ions, isotopes, pure elements and molecules (e.g., calcium, lithium), class names (e.g., steroids, fatty acids), small biochemicals, synthetic polymers. 


As shown in~\cite{tutubalina2020fair}, the CDR dataset contains a high amount of mention duplicates and overlaps between official sets. In order to obtain more realistic results, we evaluate models on preprocessed official and \emph{refined} CDR test sets from~\cite{tutubalina2020fair}. 



For the preprocessing of the clinical trial data, we use heuristic rules to split the composite mentions into separate mentions (e.g., \textit{combination of ribociclib + capecitabine} into \textit{ribociclib} and \textit{capecitabine}) by considering each mention containing ``combination'', ``combine'', ``combined'', ``plus'', ``vs'' or ``+'' as composite. 
We process all characters to lowercase forms and remove the punctuation for both mentions and synonyms.

It is assumed that each entity mention in the CDR corpus has a valid concept in the terminology, which is referred as in-KB evaluation in the entity linking task. In contrast with the CDR sets, 26\% and 1.8\% of intervention and condition mentions in the CT dataset  are not appeared in terminologies, respectively. In Section \ref{sec:outofkb}, we investigate different strategies for the out-of-KB prediction (i.e. \textit{nil} prediction) on clinical trials' texts.

\begin{table}[t]
\caption{Statistics of the datasets used in the experiments. Two sets of annotated clinical trials' fields are marked with `CT'.}
\label{tab:stats}
\begin{tabular}{|l|c|c|c|c|}
\hline
 & \textbf{CDR Disease} & \textbf{CDR Chem} & \textbf{CT Condition} & \textbf{CT Intervention} \\ \hline
domain & abstracts  & abstracts & clinical trials & clinical trials \\ 
entity type & disease & chemicals & conditions & drugs \\
terminology & MEDIC & CTD Chemicals & MeSH & in-house dict. \\
\hline
\multicolumn{5}{|c|}{entity level statistics} \\ \hline
\% numerals & 0.11\% & 7.32\% & 7.69\% & 25.3\%  \\
\% punctuation & 1.21\% & 0.07\% & 14.28\% & 24.83\% \\
avg. len & 14.88 & 11.27 & 17.92 & 21.68 \\ \hline
\multicolumn{5}{|c|}{number of pre-processed entity mentions} \\ \hline
train set & 4,182 & 5,203 & - & - \\ 
dev set & 4,244  &  5,347 & 100 &  100 \\ 
test set & 4,424 & 5,385 & 719 & 975 \\ \hline
\multicolumn{5}{|p{12cm}|}{number of pre-processed entity mentions after removal of duplicates from test set
} \\ \hline
\emph{refined} test & 657 (14.9\%) &425 (7.9\%) & 642 (78.4\%) & 846 (78.7\%) \\ \hline
\end{tabular}
\end{table}


\subsection{Baseline Methods}
We compare our proposed method with the following methods.


\paragraph{BioBERT ranking}
This is a baseline model that used the BioBERT model for encoding mention and concept representations. Each entity mention or concept name is firstly passed through BioBERT (we use the average over all outputs of BERT) and then through a mean pooling layer to yield a fixed-sized vector.
The inference task is then reduced to finding the closest concept name representation to entity mention representation in a common embedding space. We use the Euclidean distance as the distance metric. The nearest concept names are chosen as top-k concepts for entities. We use the publicly available code provided by~\cite{tutubalina2020fair} at \urlstyle{tt}\url{https://github.com/insilicomedicine/Fair-Evaluation}.




\paragraph{BioSyn}
BioSyn~\cite{sung2020biomedical} is a recent state-of-the-art model that utilizes the synonym marginalization technique and the iterative candidate retrieval. The model uses two similarity functions based on sparse and dense representations, respectively. The sparse representation encodes the morphological information of given strings via TF-IDF, the dense representation encodes the semantic information gathered from BioBERT. For reproducibility, we use the publicly available code provided by the authors at \urlstyle{tt}\url{https://github.com/dmis-lab/BioSyn}. We follow the default parameters of BioSyn as in \cite{sung2020biomedical}: the number of top candidates k is 20, the mini-batch size is 16, the learning rate is 1e-5, the dense ratio for the candidate retrieval is 0.5, 20 epochs for training.
 
\subsection{Experimental Setup}
We experiment with BioBERT$_{base}$ v1.1 with 12 heads, 12 layers, 768 hidden units per layer, and a total of 110M parameters. Epsilon, the number of positive and negative examples, and distance metric were chosen optimally on dev sets. We choose $red(\cdot)$ to be the average over all outputs of BERT. We have evaluated different epsilons starting from 0.5 up to 4.0 with 0.5 step for Euclidean distance metric, for cosine distance from 0.05 up to 0.3 with 0.05 step. These experiments have quite similar results. We have evaluated a number of positive and negative examples. For positives, we iterated over values from 15 to 35, for negatives from 5 to 15. We found that the optimal is to sample 30 positive examples and 5 negative examples per mention. For the random + parents strategy, we evaluated the number of names of concept's parents from 1 to 5. Similarly, we evaluated the number of names of concept's siblings from 1 to 5. We found that hard negative sampling (with siblings) achieves the same optima as random negative sampling. The highest metrics are achieved at $5$ concept names of the concept's parents on the CT Condition and CDR Chemical sets. The highest accuracy is achieved at $2$ names of the concept's parents on other sets. As a result, we trained the DILBERT model with Euclidean distance and the following parameters: batch size is equal to 48, learning rate was set to 1e-5, epsilon to 1.0. 

We evaluate this solution in information retrieval (IR) scenario, where the goal is to find within a dictionary of concept names and their identifiers the top-$k$ concepts for every entity mention in texts. In particular, we use the top-$k$ accuracy as an evaluation metric, following the previous works~\cite{suominen2013overview,pradhan2014semeval,wright2019normco,phan2019robust,sung2020biomedical,tutubalina2020fair}. Let Acc@k be 1 if a right CUI is retrieved at rank k, otherwise 0. All models are evaluated with Acc@1. For composite entities, we define Acc@k as 1 if each prediction for a single mention is correct.


\subsection{Out-of-KB Cases in Clinical Trials}\label{sec:outofkb}
To deal with \textit{nil} predictions in clinical trials, we apply three different strategies for the selection of a threshold value. Namely, the intervention or condition mention is considered out of KB if the nearest candidate has a larger distance than a threshold value. Our first strategy is to set the threshold equal to the minimum distance of false-positive (FP) cases. In this case, we consider a mention mapped to a concept by our model but having no appearance in the terminology. Our second strategy set the threshold to the maximum distance of true-positive (TP) cases. The third strategy uses a weighted average of the first two threshold values. The proportion of TP cases used as a weight for the first strategy's threshold, the proportion of TP cases used as a weight for the second strategy's threshold. We tested three strategies on the dev set which containing 100 randomly selected mentions and evaluated the selected threshold values on the test set. This procedure was repeated 20 times. For intervention normalization, the first strategy showed an average accuracy of 79.41 with std of 3.5; second -- accuracy of 71.77 and std of 3.5; third -- accuracy of 85.73, std of 1.3.

\subsection{Results and Discussion}\label{sec:ct}
\begin{table}[t]
\centering
\caption{Out-of-domain performance of the proposed DILBERT model and baselines in terms of Acc@1 on the \textit{refined} test set of clinical trials (CT).}\label{tab:inct_results}
\begin{tabular}{|c|c|c|c|c|}
\hline
\multirow{2}{*}{\textbf{Model}} & \multicolumn{2}{c|}{\textbf{CT Condition}} & \multicolumn{2}{c|}{\textbf{CT Intervention}} \\ 
\cline{2-5}
 & single concept & full set & single concept & full set \\ \hline
BioBERT ranking & 72.60 & 71.74 & 78.67 & 74.57 \\ \hline
BioSyn & 86.36 & -  & 86.29 & - \\ \hline
DILBERT, random sampling & 85.73 & 84.85 & 90.23 & \textbf{88.37} \\ 
DILBERT, random + 2 parents & 86.74 & 86.36 & \textbf{90.53} & 87.94 \\ 
DILBERT, random + 5 parents & \textbf{87.12} & \textbf{86.74} & 89.54 & 87.15 \\ 
DILBERT, resampling & 85.22 & 84.63 & 89.83 & 87.28 \\ 
DILBERT, resampling + 5 siblings & 84.84 & 84.26 & 89.26 & 86.23 \\ 
\hline
\end{tabular}
\begin{center}
\caption{In-domain performance of the proposed DILBERT model in terms of Acc@1 on the \textit{refined} test set of the Biocreative V CDR corpus.}\label{tab:cdr}
\begin{tabular}{|c|c|c|}
\hline
\textbf{Model} & \textbf{CDR Disease} & \textbf{CDR Chemical} \\ \hline
BioBERT ranking & 66.4 & 80.7 \\ \hline
BioSyn & 74.1 & \textbf{83.8}  \\ \hline
DILBERT, random sampling & 75.5 & 81.4 \\ 
DILBERT, random + 2 parents & 75.0 & 81.2 \\ 
DILBERT, random + 5 parents & 73.5 & 81.4 \\ 
DILBERT, resampling & \textbf{75.8} & 83.3 \\ 
DILBERT, resampling + 5 siblings & 75.3 & 82.1 \\ 
\hline
\end{tabular}
\end{center}
\end{table}

We investigate the effectiveness of transferring concept normalization from the general biomedical domain to the clinical trial domain. We trained DILBERT and BioSyn models on the CDR Disease and CDR Chemical train sets, respectively, for linking clinical conditions and interventions.  

Table \ref{tab:inct_results} presents the performance of the DILBERT models compared to BioSyn and BioBERT ranking on the datasets of clinical trials. We test the DILBERT model's transferability on two sets of interventions and conditions where each mention is associated with one concept only (see `single concept' columns). We evaluate the model on test sets with all mentions, including single concepts, composite mentions, and out-of-KB cases (see `full set' columns). In Table~\ref{tab:cdr}, we present in-domain results of models evaluated on the CDR data. In all our experiments when comparing DILBERT and BioSyn models, we use paired McNemar's test~\cite{mcnemar1947note} with a confidence level at 0.05 to measure statistical significance.

Several observations can be made based on Tables~\ref{tab:inct_results} and~\ref{tab:cdr}. First, DILBERT outperformed BioSyn and BioBERT ranking on three sets staying on par with BioSyn on the CDR Chemical test set. Adding randomly sampled positive examples from parent-child relationships gives a statistically significant improvement in 1-2\% on the CT Condition set while staying on par with random sampling on interventions. To our surprise, hard negative mining produces performance gains on one of four sets only, which includes chemicals. Second, we compare results on refined test sets with results on the CDR corpus's official test set. We observe the significant decrease of Acc@1 from 93.6\% to 75.8\% and from 95.8\% to 83.8\% for DILBERT on disease and chemical mentions, respectively. Third, DILBERT models obtained higher results on test sets with single concepts. Models achieve much higher performance for the normalization of interventions rather than conditions. The DILBERT model achieves a statistically significant improvement compared to the BioSyn model on the interventions dataset.
The error analysis on the CDR Disease set showed that models with random negative sampling incorrectly maps 39 out of 147 mentions to the correct concept's parent. We observe that some mentions are mapped to the gold concept' child for the models trained by re-sampling+siblings sampling. 

\subsubsection{Inference Time Efficiency and Deployment}

Our model uses the FAISS library~\cite{JDH17} with GPU support for fast nearest neighbor search by comparing vectors with Euclidean distance. Embeddings of all terminologies' concepts are indexed. We profiled retrieval speed on a server with Intel Xeon CPU E5-2660 2.00GHz and 256GB memory. First, we precomputed all embeddings for all concepts (500 thousand). On a single Nvidia TITAN X GPU, it takes about 7 minutes to compute all embeddings. Given that all embeddings are indexed on Nvidia TITAN X GPU using IndexFlatL2 index type. To obtain top candidates for 10 million queries, it requires approximately 3 hours. 

\section{Conclusion}
We studied the task of drug and disease normalization for clinical trials, using a newly created dataset of 500 interventional studies with 1075 intervention mentions and 819 condition mentions. We designed a triplet-based metric learning model named \texttt{DILBERT} that optimizes to pull pairs of mention and concept BioBERT representations closer than negative samples. We investigated strategies to obtain random and hard positive and negative examples using parent-child (i.e., broader-narrower) relationships between biomedical concepts. We performed experiments on in-KB and out-of-KB (\textit{nil}) linking of mentions from the scientific domain to the clinical domain in a zero-shot setting. \texttt{DILBERT} shows better transfer capabilities for disease- and drug-related mentions compared to other state-of-the-art models.
In future work, we plan to investigate taxonomy induction evaluation metrics and the normalization of protein/gene mentions. 

\paragraph{Acknowledgements}
Research on academic corpora was carried out by Z.M. and supported by RFBR, project no.~19-37-90074. 

\bibliographystyle{splncs04}
\bibliography{ml}

\end{document}